\documentclass[11pt]{article}

\usepackage[preprint]{acl}

\usepackage{times}
\usepackage{latexsym}
\usepackage{amsmath}

\usepackage[T1]{fontenc}

\usepackage[utf8]{inputenc}

\usepackage{microtype}

\usepackage{inconsolata}

\usepackage[greek,english]{babel}
\usepackage{textgreek}
\usepackage[LGR,T1]{fontenc}
\DeclareFontSubstitution{LGR}{cmr}{m}{n}
\usepackage{graphicx}
\usepackage{multirow}
\usepackage{booktabs}
\usepackage{longtable}   
\usepackage{tabularx}
\usepackage{float}
\usepackage[most]{tcolorbox}
\usepackage[dvipsnames]{xcolor}
\usepackage[table]{xcolor}
\usepackage{soul}
\newcommand{\hlc}[2][JungleGreen]{{%
    \colorlet{foo}{#1}%
    \sethlcolor{foo}\hl{#2}}%
}
\usepackage{minted}
\setminted{
  frame=single,
  bgcolor=gray!10,
  fontsize=\footnotesize,
  autogobble,
  breaklines=true,
  breakanywhere=true,
  breaksymbolleft={},
  breaksymbolright={},
  breaksymbolindentleft=0pt,
  breaksymbolindentright=0pt,
  breaksymbolsepleft=0pt,
  breaksymbolsepright=0pt
}

\newcommand{\framework}{\textsl{SAGE}} 

\newcommand{\best}[1]{\textbf{#1}}
\newcommand{\secondbest}[1]{\underline{#1}}
\usepackage{wrapfig}

\title{\framework{}: A Top-Down Bottom-Up Knowledge-Grounded User \underline{S}imulator for Multi-turn \underline{AG}ent \underline{E}valuation}

\author{
 \textbf{Ryan Shea\thanks{~~denotes equal contribution.}\textsuperscript{1}},
 \textbf{Yunan Lu$^{*}$\textsuperscript{1,2}},
 \textbf{Liang Qiu\thanks{~~The work does not relate to Liang’s work at Amazon.}},
 \textbf{Zhou Yu\textsuperscript{1,2}},
\\
 \textsuperscript{1}Department of Computer Science, Columbia University,
 \textsuperscript{2}Arklex.ai,
\\
\texttt{\{rs4235, yl4021, zy2461\}@columbia.edu}
}

\begin{document}
\maketitle
\begin{abstract}
Evaluating multi-turn interactive agents is challenging due to the need for human assessment. Evaluation with simulated users has been introduced as an alternative, however existing approaches typically model generic users and overlook the domain-specific principles required to capture realistic behavior. We propose \framework{}, a novel user \textbf{S}imulation framework for multi-turn \textbf{AG}ent \textbf{E}valuation that integrates knowledge from business contexts.  \framework{} incorporates top-down knowledge rooted in business logic, such as ideal customer profiles, grounding user behavior in realistic customer personas. We further integrate bottom-up knowledge taken from business agent infrastructure (e.g., product catalogs, FAQs, and knowledge bases), allowing the simulator to generate interactions that reflect users' information needs and expectations in a company’s target market. Through empirical evaluation, we find that this approach produces interactions that are more realistic and diverse, while also identifying up to 33\% more agent errors, highlighting its effectiveness as an evaluation tool to support bug-finding and iterative agent improvement. \footnote{Our code, documentation and demos are available at \url{https://github.com/arklexai/arksim}}
\end{abstract}

\section{Introduction}
\label{sec:introduction}

Interactive agents are increasingly deployed in high-stakes domains such as customer support, sales, and enterprise productivity \cite{muthusamy-etal-2023-towards}. As these systems become more sophisticated, evaluation has become a major bottleneck in their development. Traditional approaches rely heavily on human users, which is both costly and difficult to scale across iterative development cycles \cite{davidson2023}. Automatic evaluation methods promise greater efficiency, but existing approaches are limited. In particular, most rely on static benchmarks or narrow task-specific datasets, that fail to capture the open-ended and dynamic nature of real-world interactions \cite{yao_-bench_2024}.

Automated evaluation using user simulators has therefore emerged as a promising direction, enabling researchers and practitioners to benchmark multi-turn agents in diverse scenarios with reduced human effort. Yet most existing simulators depend heavily on human interaction data, which is often scarce, proprietary, or expensive to obtain \cite{dhole-2024-kaucus, shi-etal-2019-build}. More recent work has explored prompting-based approaches to mitigate this dependence, but these methods typically portray users as domain-general interlocutors, with goals and behaviors only loosely tied to the target application \cite{barres2025tau2benchevaluatingconversationalagents, davidson2023}. In practice, however, many real-world agents are deployed in specialized business settings where user behavior is deeply shaped by who the user is (e.g., their role, demographics, or purchasing potential) and what information they seek (e.g., company-specific product knowledge).
These dimensions are easily overlooked by simulators that fail to integrate business design principles, resulting in interactions that are less realistic and with limited utility for rigorous evaluation.


To address these limitations, we introduce \framework{}, a knowledge-grounded user simulation framework for multi-turn agent evaluation. \framework{} draws on top-down principles grounded in business logic to construct realistic user personas, incorporating attributes from ideal customer profiles (ICPs) alongside other factors that shape customer behavior. In addition, it integrates bottom-up knowledge taken from business agent infrastructure. This includes  structured and unstructured knowledge  such as product catalogs, FAQs, and knowledge bases, allowing the simulator to generate interactions that accurately reflect the knowledge and needs of relevant users of the company. By combining top-down and bottom-up knowledge, \framework{} produces more realistic and contextually relevant interactions, enabling scalable evaluation without requiring costly human efforts.

Through empirical evaluation, we demonstrate that \framework{} generates interactions that are more human-like and more diverse than existing simulation methods. Moreover, it uncovers a greater number of agent errors, underscoring its effectiveness as a tool for bug discovery and iterative agent improvement. Our contributions are as follows:

\begin{itemize}
\item We introduce \framework{}, a novel user simulation framework for multi-turn agent evaluation. \framework{} integrates top-down principles grounded in business logic and bottom-up knowledge from business agent infrastructure to generate realistic user interactions.
\item We conduct empirical evaluations illustrating that \framework{} produces interactions that are more human-like, diverse, and contextually grounded compared to existing methods.
\item We demonstrate \framework{}'s utility for identifying agent errors, finding that it identifies 25-33\% more bugs than existing methods.
\end{itemize}

\section{Related Work}
\label{sec:related}


Prior work on multi-turn agent evaluation has largely focused on creating static or task-specific benchmarks. For example, MTRAG \cite{mtRAG} introduced a human-written multi-turn RAG benchmark that evaluates end-to-end performance across four domains. $\tau$-bench \cite{yao_-bench_2024} evaluates API-augmented agents in the retail and airline domains using dynamic task-oriented dialogues between a simulated user and the target agent. CRMArena \cite{huang_crmarena-pro_2025} starts with human-written seed query templates for initial user requests and then rolls out interactions using prompt-based rules. While these methods are able to effectively assess agents in certain scenarios, their evaluations are limited to a fixed set of tasks.  \framework{} provides a flexible simulation framework that can adapt to specialized business settings, allowing it to better capture the evaluation needs of real-world agent deployment.

Previous work on user simulation methods often make heavy use of human data to generate realistic interactions. \citet{sun2025llmagentmeetsagentic} conducts a human study to collect demographic information, shopping preferences, and personality traits to build digital mapping twins. PlatoLM \cite{kong-etal-2024-platolm}, Parrot \cite{sun-etal-2024-parrot} and USP \cite{wang2025knowbettermodelinghumanlike} learn human discourse patterns directly from human-LLM interactions via supervised learning. DAUS \cite{sekulic_reliable_2024} fine-tunes an LLM on domain-specific conversation data to reduce hallucinations and maintain coherent dialogues for task-oriented dialogue. 
SimUSER \cite{bougie_simuser_2025} extracts self-consistent personas from historical user data for use in in-context learning. Although these methods can produce effective simulations, their heavily reliance on human data often limits their usability for real-world use cases where this kind of data is difficult to obtain. 

More recent work has explored prompt-based user simulators due to their cost efficiency and flexibility in domain adaptation. RecUserSim \cite{chen_recusersim_2025} introduces an LLM-based user simulator for evaluating conversational recommender systems. DuetSim \cite{luo_duetsim_2024} develops a simulator with two LLMs in tandem to ensure that the dialogue aligns with complex constraints. However, these approaches construct prompts using only generic user information without integrating the specialized knowledge needed to generate realistic interactions. As a result, they produce coherent conversations but often model generic users that are unable to capture the nuances of real-world agent interactions. Our work addresses this limitation by integrating both top-down and bottom-up knowledge, enabling simulations that are more realistic and useful for agent evaluation.

\begin{figure*}[!ht]
    \centering
    \includegraphics[scale=0.56]{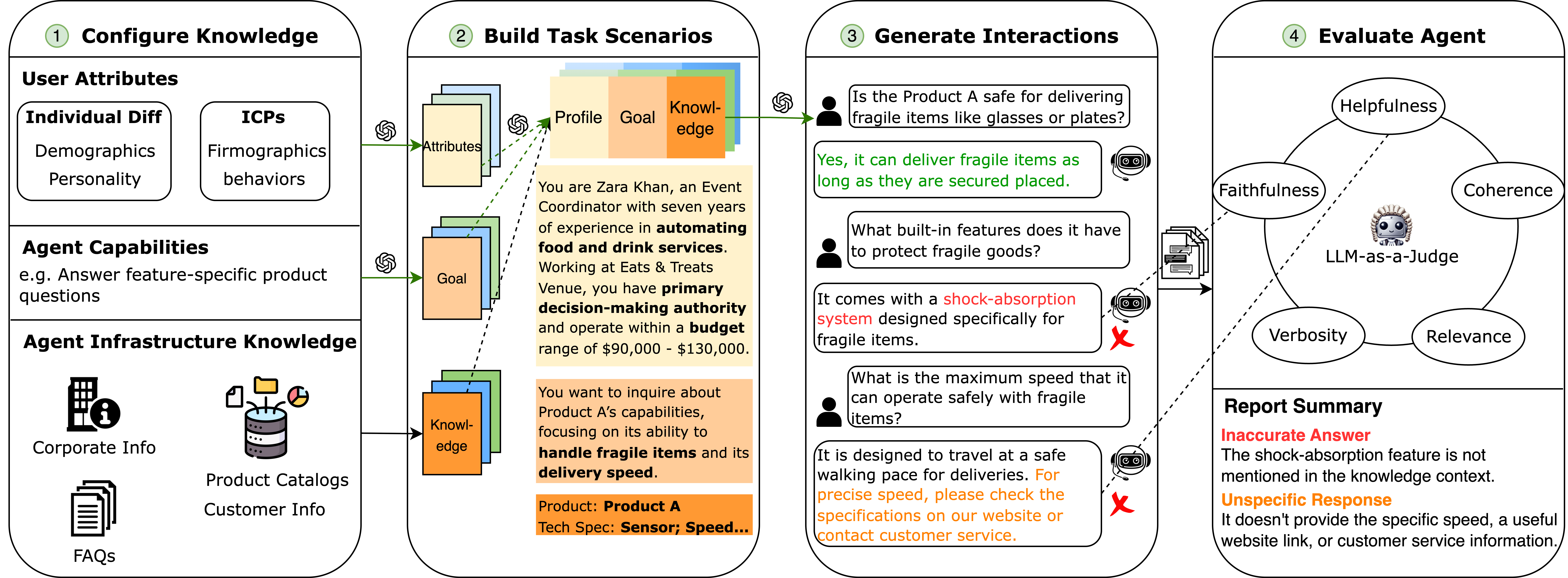}
    \caption{An overview of \framework{}, which consists of four processes: \textbf{configure knowledge}, \textbf{build task scenarios}, \textbf{generate interactions}, and \textbf{evaluate agent}. The knowledge configuration step defines the top-down (user attributes) and bottom-up (agent infrastructure knowledge) knowledge. This knowledge is leveraged to construct task scenarios which guide the user simulator in its interactions with the agent being evaluated. The resulting interactions are analyzed to identify errors in the agent's response. The complete user profile is shown in Appendix \ref{sec:app_profile} and the interaction examples are shown in Table \ref{tab:error_examples}.}
    \label{fig:pipeline}
\end{figure*}

\section{The \framework{} framework}
\label{sec:our_method}

We introduce \framework{} as a knowledge-grounded simulation framework designed for scalable multi-turn interactive agent evaluation. The process begins by configuring top-down and bottom-up knowledge including user attributes, relevant agent capabilities, and business agent resources (Section~\ref{sec:domain_knowledge}). Next, we construct task scenarios by selecting and combining these knowledge elements (Section~\ref{sec:task_scenario}). These scenarios serve as inputs to an LLM, which generates realistic multi-turn interactions with the agent being evaluated (Section~\ref{sec:gen_convo}). Finally, the generated interactions are analyzed to surface agent errors and failure cases (Section~\ref{sec:identify_bug}). An overview of the full process is provided in Figure~\ref{fig:pipeline}.

\subsection{Configure Top-down Bottom-up Knowledge}
\label{sec:domain_knowledge}
The first step in \framework{} involves configuring the knowledge base that underpins the simulated users. \framework{} integrates three complementary knowledge components. The first component, \textbf{user attributes}, defines the behavioral and personality traits that characterize each simulated user. We provide a predefined set of attributes that can be customized to align with a company’s specific requirements. The second component, \textbf{agent capabilities}, represents the set of functionalities the business aims to evaluate, supplied directly by the company. The third component, \textbf{agent infrastructure knowledge}, is provided by the company and given to the simulator to align its informational context with the agent being evaluated (Figure~\ref{fig:pipeline}).




\paragraph{User Attributes}
\label{sec:user_attributes}
The user attributes we provide are derived from top-down principles grounded in business logic. These attributes fall into two complementary categories: individual differences \cite{williamson_chapter_2018}, which include psychological and demographic factors; and customer-specific attributes, captured by ideal customer profiles. These attributes are entirely configurable and can be changed or modified to fit the company's needs.

\emph{Individual Differences} include the Big Five Personality framework \cite{BigFive_Oliver_1999} for modeling personality traits, and demographic categories following \citealp{toubia_twin-2k-500_2025}. These have been shown to be effective in predicting customer's behavior and have been used to construct simulated consumers in prior work \cite{personalitydemo_Khodabandehlou_2020, MarketSeg_Weinstein_1994}.


\emph{Ideal Customer Profiles} characterize the type of customer most valuable to a business. They include attributes such as firmographics, decision-making authority, and budget. Prior marketing studies \cite{customer_Kivist_2024, budgauth_Borgh_2020} show that these components are more predictive of customer behaviors. Marketing teams often use ICPs to formulate sales strategies and prepare for customer interactions. Incorporating ICPs into our framework enables the simulator to reflect business-relevant user characteristics and generate more realistic interactions.
See Appendix~\ref{sec:ICP} for more details.



\paragraph{Agent Capabilities} 
\label{sec:goal}
The agent capabilities are provided by the company and represent the set of functionalities that the business seeks to evaluate (Figure~\ref{fig:pipeline}). To enable the simulator to assess these capabilities, \framework{} translates each capability into a corresponding user goal using an LLM. User goals define the objective of an interaction, for instance, a user may aim to purchase a cleaning robot or inquire about the cleaning efficiency of a particular model. Creating the goals in this way anchors the objective of the user simulator while providing a systematic way to identify errors across the agent’s full range of supported tasks.


\paragraph{Agent Infrastructure Knowledge}


We also provide the simulator with bottom-up knowledge derived from agent infrastructure. The information spans both unstructured resources (e.g., webpages, FAQs) and structured data (e.g., product catalogs, inventory tables, customer records). This knowledge is provided to align the information needs and expectations of the simulated users with actual business customers. For example, a real customer will likely have read some of the company's webpages before interacting with the agent. Giving the user simulator access to the same knowledge helps align its informational context with actual users, allowing for more realistic and useful agent evaluations. This knowledge is also useful for later evaluation stages, allowing \framework{} to directly compare the information in agent responses to the ground truth infrastructure knowledge.

\subsection{Build Task Scenarios}
\label{sec:task_scenario}
Each task scenario is a natural language description composed of three components: a user goal, relevant agent infrastructure knowledge, and a user profile. We begin by choosing a user goal, then select the infrastructure knowledge most aligned with the goal. Then, we create a user profile that logically aligns with the chosen infrastructure knowledge. The final task scenario is formed by combining the components into a coherent description.


\paragraph{Goal Selection}
\label{sec:goal_selection}
To choose a user goal, we loop through the full set of goals derived in Section \ref{sec:goal}. Since the user goals are generated based on agent capabilities, looping through them provides full coverage of the tasks we want to test the agent on.

\paragraph{Agent Infrastructure Knowledge Selection}
\label{sec:business_info_selection}
Given the user goal, we adopt a hierarchical selection process to ensure that the task scenarios include diverse yet relevant infrastructure knowledge. The agent infrastructure knowledge is organized into categories (e.g., FAQs, product catalogs), each containing multiple knowledge pieces such as structured product catalogs in SQL databases or unstructured text stored in vector databases. Given a user goal (e.g., inquire about discounts or promotional offers), we first identify relevant knowledge categories by prompting an LLM. Within each selected category (e.g., FAQ), we use a clustering algorithm to find the most relevant cluster of knowledge pieces (e.g., those related to reward programs or promo codes). To encourage diversity, we sample a knowledge piece from the identified cluster that has not been previously used; if all have been used, one is randomly selected from the cluster.




\paragraph{Profile Construction}
\label{sec:profile_cons}
The user profile is generated based on the agent infrastructure knowledge and user attributes. Each user attribute consists of a key and a tuple of predefined values (e.g., purchase preference: discount oriented, economic, quality oriented). Some keys don't have predefined values since they may heavily depend on the specific use case or have too many values to enumerate (e.g., firmographics, budget). Therefore, the values are generated based on the selected infrastructure knowledge. We choose a final set of values jointly for all keys to preserve coherence among interdependent fields (e.g., job information depends on industry and company name). The final key-value pairs are used to construct the user profile, which is a natural language description of the user generated by an LLM (Figure \ref{fig:pipeline}). The related prompts are in Appendix \ref{sec:prompt_profile_construction}.

\paragraph{Task Scenario Construction}
The final task scenario is created using a structured template that combines the user goal, agent infrastructure knowledge, and user profile. The template also incorporates additional rules designed to help the model adhere more closely to the scenario and produce more human-like behavior. The constructed scenario is used as a system prompt to guide the user simulator during its interactions with the agent (see Appendix \ref{sec:prompt_profile_construction} for details).

\subsection{Generate Interactions}
\label{sec:gen_convo}

The user simulator and the agent both operate within a shared environment to ensure coherent and contextually aligned interactions. This environment integrates the interaction history, agent infrastructure knowledge, and other general world knowledge that depends on the specific use case. During interaction, both entities reference and update this environment dynamically. The user simulator is explicitly grounded on the agent infrastructure knowledge through its system prompt, whereas the agent accesses it through its internal framework, treated as a black box. The interaction history is continuously appended to both sides' context as the interaction progresses. The interaction terminates when the user emits a sentinel token (\#\#\#STOP\#\#\#) or the turn budget is reached. 




\subsection{Evaluate Agent and Identify Bugs}
\label{sec:identify_bug}
We conduct evaluations by identifying the errors that the agent makes during the generated interactions. The process begins by highlighting problematic turns using an LLM-as-a-Judge approach. Following prior work on agent evaluation \cite{zhang_crowd_2025, wang_helpsteer2_2024}, each reply is judged along five dimensions: \textbf{helpfulness}, \textbf{coherence}, \textbf{verbosity}, \textbf{relevance}, and \textbf{faithfulness}. Each dimension is given a score from 0-4 accompanied by a justification of the chosen score. \framework{}'s bottom-up knowledge integration helps enhance this scoring by offering a concrete reference point for the LLM judge to assess agent responses, rather than relying on the model’s internal knowledge.

In order to isolate agent errors, we only keep turns with a score below two. We validated this threshold through human expert verification, sampling 50 low-scoring turns (turns with scores below two) and finding that 94\% contained bugs. In contrast, among 50 high-scoring turns, only four exhibited errors. 

To give developers more concrete agent errors and actionable insights for improvement, we use an LLM to summarize the justifications from low-scoring turns into a list of unique bugs. This is done using a two-stage process. In the first stage, we prompt an LLM to generate high-level failure categories (e.g. unspecific response, inaccurate response, etc.) using the problematic turns and their score justifications. In the second stage, the output from the first stage along with the justifications are given to an LLM to generate fine-grained, unique bug statements (e.g., “did not provide weight capacity for Product A when explicitly requested”). Finally, we generate a report consisting of these bug statements along with the turn linked with each bug (see Figure \ref{fig:pipeline} for an example).

We validated the quality of the LLM-identified bugs by comparing them to bugs identified by three human experts. Across the same set of 50 low-scoring turns, the list of bugs identified by the LLM achieved an average precision of 0.73, recall of 0.74, F1 score of 0.73, and accuracy of 0.74 when compared to the human-annotated bug lists. The inter-annotator agreement among human experts, measured using Fleiss’ Kappa, was 0.85. Additional implementation details are provided in Appendix \ref{sec:app_identity_errors}, and prompt details are given in Appendix \ref{sec:prompt_agent_eval} and \ref{sec:prompt_identify_bugs}.



\section{Experiments}
\label{sec:experiments}
We conduct several experiments to evaluate the effectiveness of \framework{}'s top-down bottom-up knowledge incorporation. These experiments are designed to test the following two hypotheses:

\paragraph{H1:} Incorporating top-down knowledge such as ICPs improves the user simulator’s performance.
\paragraph{H2:} Providing bottom-up knowledge from agent infrastructure improves simulator performance.


\subsection{Evaluation Metrics}
\label{sec: user_simu_eval}
We evaluate the performance of our user simulator using the following metrics: lexical diversity, simulator fidelity, and number of agent errors identified. 

\paragraph{Lexical Diversity} To assess diversity, we utilize reference-free lexical metrics following prior work \cite{dhole-2024-kaucus}. These include Measure of Textual Lexical Diversity (MTLD), logTTR, rootTTR \cite{dhole-2024-kaucus, mccarthy2010mtld}, Hypergeometric Distribution D (HD-D) \cite{mccarthy2010mtld}, vocab size \cite{shi-etal-2019-build}, and distinct n-gram counts (for n=1,2,3) \citep{li16distinct}. 

\paragraph{Simulator Fidelity} We assess fidelity through human evaluation. We engaged two trained annotators who were tasked with rating the simulated user turns in interactions on a 3-point Likert scale across four dimensions:

\begin{itemize}
\item \textbf{Human-likeness}: How closely the simulated user’s language resembles that of a human.
\item \textbf{Coherence}: How well the simulated user’s utterances follow logically from the dialogue history.
\item \textbf{Specificity}: Whether the simulated user provides enough relevant and precise information for the system to proceed with the task.
\item \textbf{Consistency}: Whether the simulated user remains consistent in facts, preferences, constraints, and style across turns.
\end{itemize}

These dimensions follow prior work on assessing user simulator fidelity \cite{luo_duetsim_2024}. Additional details are shown in Appendix \ref{sec:app_human_eval}.

\paragraph{Error Identification}
We also assess the simulators' ability to identify agent errors. This is done by following the agent evaluation process in Section ~\ref{sec:identify_bug} and counting the number of unique bugs that were identified. A higher count of unique bugs indicates broader ability to evaluate the the agent’s capabilities and failure cases.



\begin{table*}[ht]
\centering
\begin{tabular*}{\textwidth}{@{\extracolsep{\fill}}lcccccc}
\toprule
\textbf{Method} & \textbf{MTLD} $\uparrow$ & \textbf{HDD} $\uparrow$ & \textbf{logTTR} $\uparrow$ & \textbf{rootTTR} $\uparrow$ & \textbf{Vocab Size} $\uparrow$ & \textbf{Distinct@1/2/3} $\uparrow$ \\
\midrule
\multicolumn{7}{l}{\textbf{RAG-based Sales Agent}} \\
\midrule
\emph{w/o ICP}       & 81.43 & \best{0.86} & 0.77 & 11.58 & 1011 & 0.21/0.60/0.81 \\
\emph{w/o Agent-Infra} & 80.76 & 0.85 & 0.76 & 10.65 & 973  & 0.18/0.55/0.76 \\
\framework{}              & \best{86.05} & \best{0.86} & \best{0.78} & \best{11.80} & \best{1034} & \best{0.21/0.61/0.82} \\
\midrule
\multicolumn{7}{l}{\textbf{Tool-Augmented Shopping Agent}}\\
\midrule
\emph{w/o ICP}      & 76.52 & \best{0.86} & 0.75 & 9.88  & 907  & 0.18/0.57/0.78 \\
\emph{w/o Agent-Infra} & \best{81.59} & \best{0.86} & 0.75 & 9.82  & 953  & 0.17/0.56/0.78 \\
\framework{}              & 79.97 & \best{0.86} & \best{0.76} & \best{10.37} & \best{961}  & \best{0.19/0.59/0.80} \\
\midrule \midrule
\emph{Human}         & 86.84 & 0.87 & 0.84 & 20.62 & 1360 & 0.30/0.72/0.91 \\
\bottomrule
\end{tabular*}
\caption{Lexical diversity results of \emph{w/o ICP}, \emph{w/o Agent-Infra} and \framework{} for the RAG-based Sales Agent and Tool-augmented Shopping Agent cases using
\texttt{gpt-4.1}. The best score (vs. human baseline) for each metric is shown in \textbf{bold}.}
\label{tab:diversity}
\end{table*}

\begin{table*}[htbp]
\centering
    \begin{tabular*}{\textwidth}{@{\extracolsep{\fill}}lcccc}
    \toprule
    \textbf{Method} & \textbf{Human-likeness} $\uparrow$ & \textbf{Coherence} $\uparrow$ & \textbf{Specificity} $\uparrow$ & \textbf{Consistency} $\uparrow$\\
    \midrule
    \emph{w/o ICP}       & 2.38 (0.70) & 2.75(0.49) & \best{2.80 (0.41)$^{\dagger}$} & \best{2.88 (0.33)} \\
    \emph{w/o Agent-Infra} & 2.20 (0.72) & 2.50 (0.68) & 2.53 (0.55) & \best{2.88 (0.33)} \\
    \framework{}      & \best{2.58 (0.59)$^*$} & \best{2.85 (0.36)$^*$} & 2.75 (0.44)$^{\dagger}$ & \best{2.88 (0.33)} \\
    \midrule \midrule
    \emph{Human}      & 2.90 (0.33) & 2.85 (0.36) & 2.85 (0.42) & 2.90 (0.30) \\
    \bottomrule
    \end{tabular*}

\caption{Human evaluation results of \emph{w/o ICP}, \emph{w/o Agent-Infra}, \framework{} and \emph{Human} interactions with the RAG-based Sales Agent. Each condition is rated on a 3-point Likert scale across four dimensions. The best score (vs. human baseline) for each metric is shown in \textbf{bold}, standard deviations are in parenthesis. Statistically significant improvements (independent two-sample t-test, $p < 0.05$) over both baselines are marked with $^*$, improvements over one baseline are marked with $^{\dagger}$.} 
\label{tab:human_eval}
\vspace{-0.5em}
\end{table*}


\begin{table}
\centering
\scalebox{1}{
    \begin{tabularx}{\columnwidth}{@{\extracolsep{\fill}} p{2.42cm}@{\hspace{5pt}} c c}
    \toprule
    \multirow{2}{*} & \multicolumn{2}{c}{Number of unique bugs} \\
    \cmidrule(lr){2-3}
    \textbf{Method}&  \textbf{RAG-based} & \textbf{Tool-Augmented}\\
    \midrule
    \emph{w/o ICP} & 29.0 (1.78) & 14.4 (2.10)\\
    \emph{w/o Agent-Infra} & 13.6 (1.02) & 13.0 (1.02)\\
    \framework{} & \best{38.6 (1.24)} & \best{18.0 (1.62)}\\
    \midrule \midrule
    \emph{Human} & 14.0 (2.60) & -- \\
    \bottomrule
    \end{tabularx}
}
\caption{The average counts of unique bugs found over five runs with \emph{w/o ICP}, \emph{w/o Agent-Infra}, \framework{}, and \emph{Human} interactions. The highest counts are shown in \textbf{bold}, standard deviations are in parenthesis.}
\label{tab:unique_bugs}
\vspace{-1.6em}
\end{table}

\subsection{Experimental Setup}
\label{sec:exp_setup}

\paragraph{Settings} We conduct our experiments by comparing \framework{} against two different conditions: 

 \emph{w/o ICP}: In this condition, the user simulator is given all knowledge except for top-down knowledge with respect to ideal customer profiles. Instead it has to rely exclusively on individual differences (Section \ref{sec:user_attributes}) and bottom-up knowledge to generate realistic interactions. This approach follows prior methods of constructing prompt based user simulators such as the ones in $\tau$-bench \cite{yao_-bench_2024} and CRMArena \cite{huang_crmarena-pro_2025}.
 
 \emph{w/o Agent-Infra}: In this setting, the user simulator is given all knowledge except for bottom-up knowledge from the company's agent infrastructure. It is only provided with high-level corporate overviews and top-down knowledge. This approach mirrors existing methods of creating persona-based user simulators such as DuetSim \cite{ahmad_simulating_2025, luo_duetsim_2024}.

\paragraph{Use Cases}
We conduct our experiments on two customer-facing agents: (i) a production-deployed RAG-based agent, and (ii) a tool-augmented shopping agent deployed on the Shopify platform.

\emph{RAG-based Sales Agent}
We simulate interactions with a production-deployed RAG agent used by a robotics company. The agent is designed to provide answers to customer inquiries about the company's products and services using technical product catalogs, FAQs, and corporate information. 


\emph{Tool-augmented Shopping Agent}
We also generate interactions with a shopping agent deployed on the Shopify platform. The agent integrates 13 APIs, such as \texttt{get\_recommendation}, \texttt{get\_user\_info}, \texttt{get\_product}, \texttt{return\_item}, and \texttt{cancel\_order}, alongside FAQs and store policies.

These settings are designed to test the two hypotheses outlined previously. We test each setting using different LLMs from both open-source and commercial model families: \texttt{gpt-4.1} \cite{openai-gpt41}, \texttt{gpt-5} \cite{openai-gpt5}, \texttt{gemini-2.5-flash} \cite{gemini}, \texttt{Qwen2.5-7B-Instruct} \cite{qwen2.5} and \texttt{Llama-3.3-70B-Instruct} \cite{llama3.3} to show \framework{}'s ability to generalize across different models. The results with \texttt{gpt-4.1} are in Section \ref{sec:results}, evaluations for other models are in Appendix \ref{sec:app_auto_eval}.

\paragraph{Human-Agent Interactions Data}
To ground our evaluation in real-world user behavior, we collect a dataset of real customer interactions with the production-deployed RAG-based sales agent. The agent is designed to answer product-related questions and maximize lead conversion rate, which is the proportion of users who provide contact information among all users who interact with the agent. We log all human–agent interactions collected between July 2024 and July 2025. To ensure comparability with our simulation setting, we retain only English-language interactions.

To focus the dataset on target customer engagement rather than noise or adversarial usage, we postprocess the raw dataset using outcome-driven filtering. Specifically, we analyze lead conversion rates as a function of conversation length and observe that interactions spanning 3 to 15 turns achieve conversion rates above 20\%. In contrast, extremely short conversations (1–2 turns, typically limited to greetings) and excessively long conversations (>15 turns, often driven by off-topic queries) exhibit substantially lower conversion rates. Based on this analysis, we retain conversations within the $[3, 15]$ turn range for downstream comparison with simulated interactions.

\begin{table*}
    \centering
    \small
    \begin{tabular*}{\textwidth}{@{\extracolsep{\fill}} p{0.3\textwidth} p{0.3\textwidth} p{0.3\textwidth}}
    \toprule
    \textbf{w/o ICP} & \textbf{\framework{}} &\textbf{w/o Agent-Infra}\\
    \midrule
    \textbf{User}: \textcolor{magenta}{I'm interested in the \hlc[SkyBlue]{Product\_A} delivery robot. How does it compare to other delivery solutions?} \newline
    \textbf{Agent}: The Product\_A delivery robot is \ldots For more details, you can explore our website. Which model(s) are you interested in? \newline
    \textbf{User}: \textcolor{magenta}{I'm mainly interested in delivery solutions and comparing Product\_A. Can you tell me more about its benefits over other robots?} \newline
    \textbf{Agent}: The Product\_A delivery robot excels in indoor environments like restaurants and cafes, \ldots \newline
    \textbf{User}: Can it handle \hlc[SkyBlue]{different types of deliveries} efficiently? \newline
    \textbf{Agent}: Yes, it can efficiently handle various types of deliveries. \ldots To speed up your process, would you like to provide your business name? \newline
    \textbf{User}: I'm currently exploring options. Thanks for the info!
    &
    \textbf{user}: hey, how does \hlc[SkyBlue]{Product\_A} compare with other delivery robots? \hlc[SpringGreen]{need it for agriculture operations}. \newline
    \textbf{Agent}: Product\_A is designed for efficient delivery operations. \ldots Which model(s) are you interested in? \newline
    \textbf{User}: i'm trying to evaluate if Product\_A is suitable for \hlc[SpringGreen]{outdoor environments in farming}. what's its durability like? \newline
    \textbf{Agent}: \textcolor{orange}{Product\_A is designed for robust outdoor use, making it suitable for agricultural operations.} \ldots Do you want to rent or buy our robots? \newline
    \textbf{User}: \hlc[SkyBlue]{how long does it take to charge?} i'm considering all options. \newline
    \textbf{Agent}: It takes approximately \textcolor{orange}{3 hours} to fully charge from 0 to 100\%. \ldots To expedite your rental process, could you please provide your business name? \newline
    \textbf{User}: \hlc[SpringGreen]{it's AgriFuture Technologies}. not renting yet, just exploring options.
    &
    \textbf{User}: \textcolor{magenta}{How do the Product\_B and Product\_C compare} \hlc[SpringGreen]{for farm work}? \newline
    \textbf{Agent}: The Product\_B is versatile and can be used for multiple purposes. \ldots \newline
    \textbf{User}: What about Product\_A robot? \newline
    \textbf{Agent}: The Product\_A robot is a delivery robot designed for transporting items. \ldots Which model(s) are you interested in? \newline
    \textbf{User}: Delivery bot like Product\_A I think. \hlc[SpringGreen]{Can it handle outdoor conditions?} \newline
    \textbf{Agent}: Currently, our products are primarily designed for indoor business environments, but they may also be suitable for outdoor conditions. \ldots \newline
    \textbf{User}: Got it. Thanks for the info. \\
    \bottomrule
    \end{tabular*}
    \caption{Interaction samples from our human evaluation under three conditions: \framework{} (middle), \emph{w/o ICP} (left), and \emph{w/o Agent-Infra} (right) with the same user's goal: compare products. Text highlighted in \hlc[SkyBlue]{blue} reflects generated content tied to infrastructure knowledge, while text in \hlc[SpringGreen]{green} reflects content consistent with ICP attributes. Agent responses marked in \textcolor{orange}{orange} indicate faithfulness errors. On the left, user queries in \textcolor{magenta}{magenta} indicate repetitive and generic responses. On the right, the user queries in \textcolor{magenta}{magenta} are ones which are irrelevant to product attributes. Product\_A, Product\_B and Product\_C are placeholders used to mask the actual product names. The complete interaction examples are provided in Table \ref{tab:error_examples_full}.}
    \label{tab:error_examples}
\end{table*}


\section{Results}
\label{sec:results}

The results of our experiments are in the following sections. On average, \framework{} performs the best across all evaluations, supporting both of the hypotheses outlined in Section \ref{sec:experiments}. This showcases the importance of including both top-down ICP knowledge and bottom-up agent infrastructure knowledge to generate high quality simulated interactions.

\subsection{Automatic Evaluation}

We evaluated lexical diversity and error identification rate for simulated and real human-agent interactions under our different settings. For simulated interactions, we generated 150 interactions per condition and use case. For the real human-agent interactions, we sampled 150 conversations from the post-processed dataset. Lexical diversity was computed directly from these interactions. To assess error identification, we identified the number of unique agent bugs in the interactions using the process in Section \ref{sec:identify_bug}. This process was repeated five times to control for any variation in the bug-finding process. The final results report the average number of bugs discovered.

Our lexical diversity evaluation is in Table \ref{tab:diversity}. \framework{} produces interactions that are closer to real human interactions across all diversity metrics with the exception of MTLD score in the Tool-augmented Shopping Agent case. The \emph{w/o ICP} condition does slightly better than the \emph{w/o Agent-Infra} condition in the RAG-based Sales Agent case, although both perform similarly in the Tool-Augmented Shopping Agent case.

Table \ref{tab:unique_bugs} shows the outcome of our error identification evaluation. \framework{} identifies the most errors followed by the \emph{w/o ICP} condition. Excluding bottom-up knowledge leads to the largest decrease in errors found, especially in the case with the RAG-based agent. This is likely due to the fact that bottom-up knowledge allows the simulator to ask more specific questions, which helps directly target the agent's knowledge retrieval capabilities. However, there are fewer bugs present in the human interactions which suggests that user simulation methods may actually be more effective at identifying and correcting agent errors compared to relying on human interactions.

\subsection{Human Evaluation}

Our human evaluation of simulator fidelity is based on 120 annotated interactions (40 per condition) between the simulators and the RAG-based Sales Agent. In addition, we sample 40 conversations from real customer interactions with the same agent to serve as the reference distribution.

The results of the evaluation are in Table \ref{tab:human_eval}. The interactions generated by \framework{} are significantly better than the other conditions and are closer to real human-agent interactions in terms of human-likeness and coherence. The simulation method without ICPs performs slightly better in terms of specificity, suggesting that only bottom-up knowledge is essential to achieve high performance. All conditions perform similarly with respect to consistency, indicating that knowledge integration may not be necessary to perform well in this dimension. The \emph{w/o ICP} condition tends to outperform the \emph{w/o Agent-Infra} condition, implying that agent-infrastructure knowledge may be more important than ICP knowledge for generating high-fidelity simulated interactions in the case of RAG agents.

\subsection{Case Study}
\label{sec:error_analysis}
To further illustrate the differences among simulation methods, we conducted a case study using three interaction samples drawn from our human evaluation. The samples were generated by \framework{}, \emph{w/o ICP}, and \emph{w/o Agent-Infra} (see Table~\ref{tab:error_examples}). Each method was applied to the same goal, comparing company products, enabling a direct comparison of simulator fidelity and evaluation effectiveness.

\framework{} was the only method that identified agent errors in the interaction. Both of these were faithfulness errors where the agent failed to accurately retrieve information from its knowledge base. Specifically, the agent incorrectly stated that Product\_A was “designed for robust outdoor use” and required only 3 hours to charge, when in fact it is intended primarily for indoor use and requires 5.5 hours. The first error was triggered after the simulator asked if Product\_A was suitable for outdoor use in farming, a query informed by its top-down ICP knowledge. The second arose when the simulator inquired about charging time, which is based on its bottom-up agent infrastructure knowledge. Together, these cases illustrate how both top-down and bottom-up knowledge integration contribute to identifying different agent errors.

We observe that the simulator \emph{w/o ICP} often produces generic or repetitive responses (see Table \ref{tab:error_examples}, first column), resulting in less diverse conversational trajectories and reducing its effectiveness as an evaluation tool. This limitation appears to stem from the use of a more generic persona that lacks the fine-grained attributes provided by ICPs. Such attributes are essential for modeling customer behavior and enabling realistic, contextually useful interactions in business settings \cite{customer_Kivist_2024}. As a result, the \emph{w/o ICP} simulator tends to underestimate the complexity of real-world user behavior, leading to less useful interactions.

The simulator \emph{w/o Agent-Infra} frequently produces interactions that are misaligned with the company’s informational context. For example, as shown in the third column of Table \ref{tab:error_examples}, it asks the agent to compare Products B and C for farm work, despite Product\_B being designed for serving food and drinks in a restaurant setting and thus irrelevant to farming. Without access to agent infrastructure, the simulator generates less realistic conversations and tends to default to generic queries, which limits its ability to rigorously evaluate agent responses or test their information retrieval capabilities.

\begin{figure}
    \centering
    \includegraphics[width=\columnwidth]{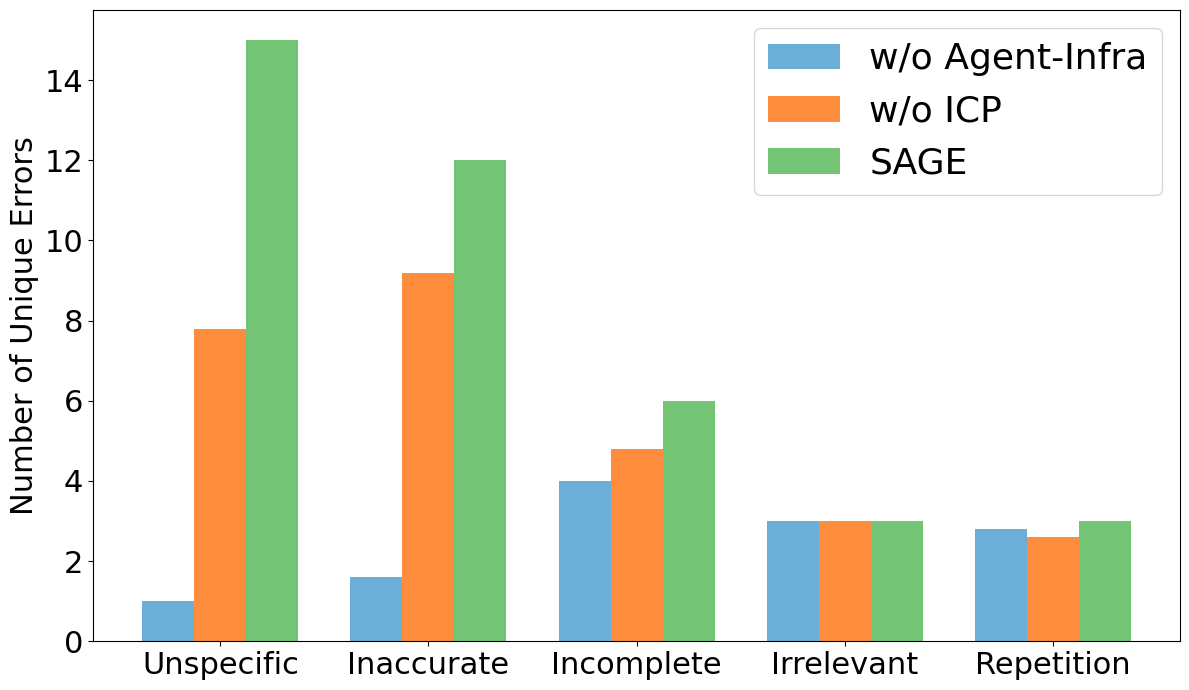}
    \caption{Number of unique errors across high-level error categories for the RAG-based Sales Agent under \emph{w/o ICP}, \emph{w/o Agent-Infra}, and \framework{} settings.}
    \label{fig:unique_bugs_category}
    \vspace{-1em}
\end{figure}


\subsection{Error Analysis}
\label{sec:error_categories}

Figure \ref{fig:unique_bugs_category} shows the distribution of high-level error categories identified by \framework{}, \emph{w/o ICP}, and \emph{w/o Agent-Infra} during our evaluation of the RAG-based Sales Agent. The most frequent issues found by \framework{} fall under the categories of ``Unspecific Response'' and ``Inaccurate Answer,'' which aligns with the relevant functionalities of a RAG-based system. This demonstrates that \framework{} effectively captures the agent’s most relevant failure modes.  The \emph{w/o Agent-Infra} struggles to find errors in these categories, highlighting the importance of incorporating agent infrastructure knowledge to induce and identify errors related to information retrieval. We also observe that the vast majority of errors emerge after the first turn, underscoring the need for multi-turn evaluation when assessing agent behavior.

\section{Conclusion}
\label{sec:conclusion}
User simulators have been proposed as a scalable alternative for evaluating multi-turn interactive agents. However, existing approaches often rely on static benchmarks or model generic users. To address this, we proposed \framework{}, a knowledge-grounded user simulation framework for multi-turn agent evaluation. \framework{} combines top-down principles derived from business logic with bottom-up knowledge from agent infrastructure to improve simulation quality. We evaluated our framework using automatic metrics and human assessments, finding that \framework{} produces interactions that are more diverse, realistic, and useful for agent evaluation.

\section{Limitations}
\label{sec:limitations}
One limitation of the \framework{} framework is that its predefined user attributes are derived from business logic, which mostly relates to customer-facing agents. While these attributes can be changed as needed to adapt to different use cases, developers creating internal-facing agents for education, invoicing, or other back-office tasks may need to customize the attributes for their specific needs.


Our current task scenarios only contain a single user goal and one piece of knowledge selected from agent infrastructure knowledge. Although real customer-agent sessions are often short, users may still switch to new topics without starting a new session. Future work should incorporate multiple goals and multiple knowledge sources within a single session to capture more complex interaction and better reflect real-world conversation patterns.

Finally, our user simulator evaluation focuses on a cold-start scenario without a direct comparison to real user interactions. The automatic diversity evaluation is limited to reference-free lexical measures, and the fidelity assessment relies on human judgments. Although this setup reflects realistic development conditions where a new agent lacks historical interaction data, future work could benefit from using real-world logs as evaluation references once production-level interaction data becomes available.

\section{Ethics Statement}
This work studies LLM-based user simulation for agent evaluation and raises considerations regarding fairness and bias in the representation of user attributes. Generating user attributes using LLMs rather than sampling from real user distributions, may lead to over-representation of majority groups and under-representation of minority groups \cite{lahoti_improving_2023, kotek_gender_2023}, potentially skewing evaluation results.

Prior work has also shown that LLMs tend to model majority-group behaviors more faithfully than those of minority groups, potentially resulting in systematic performance gaps when simulators are used to assess agent behavior across diverse user populations \cite{durmus_towards_2024, santurkar_whose_2023}. Moreover, when explicitly conditioned on attributes associated with minority groups, simulators may produce stereotyped representations, reinforcing harmful biases \cite{cheng_marked_2023, nadeem_stereoset_2020}.

Our framework does not claim to resolve these challenges. Instead, it should be viewed as an evaluation tool whose fidelity depends on the quality and balance of the underlying attribute specifications and knowledge sources. In this work, user attributes are used primarily to control and stress-test agent behavior rather than to faithfully reproduce population-level distributions.

\section{Acknowledgments}
We would like to thank Xinyang Wang, Xuanming Zhang, and Nikita Rajaneesh for their help in conducting human evaluations. We would also like to thank Richtech Robotics for providing Human-Agent interaction data for evaluation.

\bibliography{custom}

\clearpage
\appendix
\newpage

\section{Implementation Details}
\label{sec:implementation_details}


\subsection{Identity Errors}
\label{sec:app_identity_errors}
We evaluate agent performance through systematic error identification. The process begins by scoring each agent response at the turn level using an LLM-as-a-Judge approach. Specifically, we employ the \texttt{gpt-4.1} model (\texttt{gpt-4.1-2025-04-14}) with \texttt{temperature=0.1} as the evaluation model. Turns that receive a score below a predefined threshold (e.g., 2 on a 5-point Likert scale) are flagged for further inspection. The threshold is determined through human alignment: annotators review the LLM-provided scores and justifications to establish which values should be considered indicative of an error. This alignment step is critical because different evaluation models exhibit different scoring biases. For example, we find \texttt{gpt-5} tends to assign lower helpfulness scores on average than \texttt{gpt-4.1}. After human verification, we choose to use 2 as the threshold for result evaluated by \texttt{gpt-4.1} model to retain the problematic turns. 

Table~\ref{tab:agent_eval} reports the LLM-as-a-Judge evaluation results on interactions generated by \texttt{gpt-4.1} with \texttt{temperature=0.1}. Each entry indicates the proportion of turns, across 150 conversations, that received a score below 2. A higher percentage reflects a greater share of turns exhibiting potential issues. The results suggest that \emph{w/o Agent-Infra} produces more problematic turns. However, many of these problematic turns contain duplicate errors, meaning that the \emph{w/o Agent-Infra} condition doesn't actually provide a wider coverage of agent failure modes. We identify the number of unique bugs in order to properly assess the simulators' ability to identify diverse agent errors. Additionally, the bug statements we provide pinpoint concrete failure cases and give actionable insights which are not captured by the LLM judge's justifications. The detailed discussion for interactions generated by other models (\texttt{gpt-5} and \texttt{gemini-2.5-flash}) are shown in Appendix \ref{sec:app_auto_eval}.



\begin{table*}
\centering
\begin{tabular*}{\textwidth}{@{\extracolsep{\fill}}p{2.42cm}@{\hspace{4pt}}ccccc}
\toprule
 \textbf{Method} & \textbf{Helpfulness}(\%) & \textbf{Coherence}(\%) & \textbf{Verbosity}(\%) & \textbf{Relevance}(\%) & \textbf{Faithfulness}(\%) \\
\midrule
\multicolumn{6}{l}{\textbf{RAG-based Sales Agent}} \\
\midrule
\emph{w/o ICP}       & 5.18 & 0.00 & 7.14  & 2.14 & \secondbest{7.71} \\
\emph{w/o Agent-Infra} & \best{8.48} & \best{3.78} & \best{13.03} & \best{6.67} & 6.06 \\
\framework{}              & \secondbest{6.37} & 0.00 & \secondbest{7.29}  & \secondbest{3.64} & \best{8.56} \\
\midrule
\multicolumn{6}{l}{\textbf{Tool-Augmented Shopping Agent}}\\
\midrule
\emph{w/o ICP}       & 7.35  & \secondbest{4.68} & 15.05 & 8.69  & 8.19 \\
\emph{w/o Agent-Infra} & \best{11.67} & 4.66 & \best{22.43} & \best{12.71}  & \secondbest{8.43} \\
\framework{}              & \secondbest{11.13} & \best{4.73} & \secondbest{15.85} & \secondbest{10.82} & \best{9.76} \\
\bottomrule
\end{tabular*}
\caption{The LLM-as-a-Judge evaluation results for the \emph{w/o ICP}, \emph{w/o Agent-Infra}, and \framework{} using \texttt{gpt-4.1} with temperature \texttt{0.1} as user simulator model. Each value denotes the proportion of turns (across 150 conversations) rated below 2 on a 5-point Likert scale. The highest values are shown in \textbf{bold}, and the second highest are \underline{underlined}.}
\label{tab:agent_eval}
\end{table*}

\section{ICP}
\label{sec:ICP}
We distinguish ICP attributes by Business-to-Business (B2B) and Business-to-Consumer (B2C) contexts, since they differ fundamentally in decision-making processes and relevant information. B2B interactions are typically collective and process-driven, where attributes including firmographics (e.g., industry, company size) or authority are central. While B2C interactions are shaped by individual decision-making styles, such as purchasing preferences, spending behavior \cite{spending_behavior_2008} and loyalty levels. 


\begin{itemize}
    \item Business-to-Business (B2B) Attributes : Following prior work \cite{customer_Kivist_2024, budgauth_Borgh_2020}, we identify the most common B2B attributes (with predefined values listed in parentheses): \emph{customer type} (new prospect; returning customer), \emph{discovery channel} (search engine; third party website; friend recommendation; social media), \emph{budget}, \emph{decision-making authority} (primary; secondary), \emph{deal stage} (new leads; demo schedule; after demo; decision make; contract sent; contract signed; paid), \emph{industry}, \emph{company name}, \emph{job information}, and \emph{company size}. Some categories (e.g., authority, customer type) admit small, stable value sets; others (e.g., company name, industry) are generated dynamically.
    \item Business-to-Consumer (B2C) Attributes: For B2C we reuse the transferable fields from the B2B attributes (customer type, discovery channel, budget) and add other attributes that specific for B2C scenarios. \emph{decision-making style} (rational; intuitive), \emph{purchasing preference} (discount‑oriented; value‑oriented; quality‑oriented), \emph{spending behavior} (trouble spending money; some trouble spending mony; no trouble spending money; some trouble limiting spending; over spending) \cite{spending_behavior_2008}, and \emph{loyalty level} (loyal; neutral; disloyal) \cite{sun2025llmagentmeetsagentic}.
\end{itemize}

\begin{table*}
    \centering
    \begin{tabular}{ll}
    \toprule
      \textbf{Stages}   &  \textbf{Agent Capabilities}\\
      \midrule
      Need Recognition  &  Assist purchase planning based on scenario; \\
      \midrule
      \multirow{2}{*}{Information Search} & Recommend products within defined categories and constraints; \\
      & Answer feature-specific product questions; \\
      \midrule
      Evaluation of Alternatives & Provide comparisons of products; \\
      \midrule
      \multirow{2}{*}{Purchase Decision} & Answer queries about delivery; \\
      & Answer queries about discount and promotions offers; \\
      \midrule
      \multirow{3}{*}{Post-Purchase Behavior} & Handle queries about order tracking and status updates; \\
      & Handle reports about product issues or defects; \\
      & Handle requests for order refund or returns; \\
    \bottomrule
    \end{tabular}
    \caption{Heuristics of agent capabilities for the shopping agent.}
    \label{tab:agent_cap}
\end{table*}

\begin{figure*}
\begin{tcolorbox}[
  breakable,
  enhanced,
  colback=gray!5!white,
  colframe=gray!60!black,
]
\small{
You are \textbf{Zara Khan} \textcolor{Green}{[name]} who is \textbf{conscientious, extroverted, antagonistic, neurotic, and closed to experience} \textcolor{Green}{[personality]}. As a \textbf{32-year-old} \textcolor{Green}{[age]} \textbf{male} \textcolor{Green}{[sex]}, you are a \textbf{college graduate with some postgraduate education} \textcolor{Green}{[education]} and call \textbf{Karachi, Pakistan, your home} \textcolor{Green}{[citizenship, location]}. You bring your skills to bear as \textbf{an Event Coordinator with seven years of experience} \textcolor{Green}{[job information]} in \textbf{automating food and drink services} \textcolor{Green}{[business type]} at large-scale events. Working at \textbf{Eats \& Treats Venue} \textcolor{Green}{[company name]}, a hospitality company with \textbf{11-50 employees} \textcolor{Green}{[company size]}, you have \textbf{primary decision-making authority} \textcolor{Green}{[decision-making authority]} and operate within a budget range of \textbf{\$90,000 - \$130,000} \textcolor{Green}{[budget]}. Your personal life is marked by being \textbf{a divorced father} \textcolor{Green}{[marital status]} in \textbf{a household of more than four} \textcolor{Green}{[household]}, and you navigate your social and professional spheres through \textbf{a democratic lens} \textcolor{Green}{[political affiliation]}. Despite being rooted in your traditions and having a firm worldview, you maintain a professional curiosity for innovative approaches, particularly those that can elevate the events you coordinate. Having discovered <Company> through \textbf{a friend's recommendation} \textcolor{Green}{[discovery type]}, you're interested in exploring their service robotics solutions, recognizing the potential benefits they can offer in seamlessly integrating into your venue's operations. \textbf{As a new prospect} \textcolor{Green}{[customer type]} \textbf{at the new leads stage} \textcolor{Green}{[deal stage]}, you have a keen eye on how technologies can streamline your services, and you're poised to enhance your venue's offerings in the competitive \textbf{hospitality industry} \textcolor{Green}{[business type]}.
}
\end{tcolorbox}
\caption{A user profile example}
\label{fig:user_profile}
\end{figure*}

\section{Agent Capabilities}
\label{sec:app_agent_cap}
Agent capabilities capture the concrete functionalities an interactive agent is designed to perform. In the e-commerce, these functions broadly serve the goal of streamlining consumer decision-making and enabling efficient customer journeys. Following established consumer decision-making models \cite{decision-making-2019, decision-making-2021}, we categorize the capabilities into five categories: \emph{Need Recognition}, \emph{Information Search}, \emph{Evaluation of Alternatives}, \emph{Purchase Decision}, and \emph{Post-Purchase Behavior}. Inspired by real-world systems such as Amazon Rufus and Walmart Sparky, we further refine these capabilities into concrete, testable instructions using a Verb–Object–Prepositional Phrase (constraint) structure \cite{requirements-engineering-2009}. The full heuristics are summarized in Table~\ref{tab:agent_cap}.

\section{Profile Sample}
\label{sec:app_profile}
The complete user profile in Figure \ref{fig:pipeline} is shown in Figure \ref{fig:user_profile}. The corresponding attribute categories are in brackets. The sensitive information about the company is masked through human review.

\section{Human Evaluation Details}
\label{sec:app_human_eval}
For human evaluation, we recruited two graduate students as annotators. Both had prior experience with developing interactive agents, which qualified them to perform the task. The annotators conducted the evaluation collaboratively, discussing each case until they reached a consensus score. Detailed annotation guidelines are provided below. Our data collection protocols are IRB approved.

\begin{tcolorbox}[
  breakable,
  enhanced,
  colback=gray!5!white,
  colframe=gray!60!black,
  left=5pt, 
  right=5pt, 
]
\small{
\textbf{1. Human-Likeness} \\
\textbf{Definition}: How closely the simulated user’s language resembles that of a human. \\
\textbf{Scale (1–3)}: \\
1 = Clearly artificial, awkward, or robotic phrasing. \\
2 = Somewhat human-like with occasional awkwardness or unnatural phrasing. \\
3 = Highly natural; indistinguishable from a real user. \\
\\
\textbf{2. Coherence} \\
\textbf{Definition}: How well the simulated user’s utterances follow logically from the dialogue history. \\
\textbf{Scale (1–3)}: \\
1 = Often off-topic or logically inconsistent. \\
2 = Mostly coherent with occasional minor lapses. \\
3 = Fully coherent throughout. \\
\\
\textbf{3. Specificity} \\
\textbf{Definition}: Whether the simulated user provides enough relevant and precise information for the system to proceed with the task. \\
\textbf{Scale (1–3)}: \\
1 = Missing essential details or providing irrelevant content. \\
2 = Provides partial or vague information. \\
3 = Provides all necessary, relevant, and concise information. \\
\\
\textbf{4. Consistency} \\
\textbf{Definition}: Whether the simulated user remains consistent in facts, preferences, constraints, and style across turns. \\
\textbf{Scale (1–3)}: \\
1 = Contradicts previous statements or changes facts/preferences without reason. \\
2 = Mostly consistent with minor contradictions or shifts. \\
3 = Fully consistent across the dialogue.
}
\end{tcolorbox}

\begin{table*}
\centering
\begin{tabular*}{\textwidth}{@{\extracolsep{\fill}}lccccccc}
\toprule
\textbf{Method} & \textbf{MTLD}$\uparrow$ & \textbf{HDD}$\uparrow$ & \textbf{logTTR}$\uparrow$ & \textbf{rootTTR}$\uparrow$ & \textbf{Vocab Size}$\uparrow$ & \textbf{Distinct@1/2/3}$\uparrow$ \\
\midrule
\multicolumn{7}{l}{\textbf{RAG-based Sales Agent}} \\
\midrule
w/o ICP       & 84.17 & \best{0.86} & 0.77 & 11.87 & 1076 & \best{0.20/0.60/0.81} \\
w/o Agent-Infra & \best{94.01} & \best{0.86} & 0.76 & 10.56 & 1053  & 0.16/0.54/0.76 \\
\framework{}  & 88.01 & \best{0.86} & \best{0.78} & \best{12.26} & \best{1150} & \best{0.20/0.60/0.81} \\
\midrule
\multicolumn{7}{l}{\textbf{Tool-Augmented Shopping Agent}}\\
\midrule
\emph{w/o ICP}       & 66.97 & 0.85 & 0.75 & 10.07 & 947 & \best{0.18/0.58/0.79}  \\
\emph{w/o Agent-Infra} & \best{77.89} & \best{0.86} & 0.75 & 9.81 & 932  & 0.17/0.57/0.78 \\
\framework{}  & 72.57 & \best{0.86} & \best{0.76} & \best{10.45} & \best{951} & \best{0.18/0.58/0.79} \\
\bottomrule
\end{tabular*}
\caption{Lexical diversity results of \emph{w/o ICP}, \emph{w/o Agent-Infra} and \framework{} for the RAG-based Sales Agent and Tool-augmented Shopping Agent cases using
\texttt{gpt-5}. The best score for each metric is shown in \textbf{bold}.}
\label{tab:app_diversity_gpt5}
\end{table*}

\begin{table*}
\centering
\begin{tabular*}{\textwidth}{@{\extracolsep{\fill}}lcccccc}
\toprule
\textbf{Method} & \textbf{MTLD}$\uparrow$ & \textbf{HDD}$\uparrow$ & \textbf{logTTR}$\uparrow$ & \textbf{rootTTR}$\uparrow$ & \textbf{Vocab Size}$\uparrow$ & \textbf{Distinct@1/2/3}$\uparrow$ \\
\midrule
\multicolumn{7}{l}{\textbf{RAG-based Sales Agent}} \\
\midrule
w/o ICP      & 91.14 & 0.87 & 0.77 & 15.41  &  2241 & 0.15/0.53/0.77 \\
w/o Agent-Infra & 101.41 & 0.88 & 0.76 & 15.97  & 2708  & 0.13/0.50/0.76 \\
\framework{}  & \best{105.08} & \best{0.89} & \best{0.78} & \best{17.55} & \best{2765}  & \best{0.16/0.55/0.79} \\
\midrule
\multicolumn{7}{l}{\textbf{Tool-Augmented Shopping Agent}}\\
\midrule
\emph{w/o ICP}   & 74.94  & \best{0.87} & \best{0.75} & \best{12.88} & 1755 & 0.14/0.48/0.73   \\
\emph{w/o Agent-Infra} & 76.58 & 0.86 & \best{0.75} & 12.58 & 1743 & 0.13/0.48/0.73 \\
\framework{}  & \best{77.52} & \best{0.87} & \best{0.75} & \best{12.88} & \best{1851} & \best{0.14/0.48/0.74} \\
\bottomrule
\end{tabular*}
\caption{Lexical diversity results of \emph{w/o ICP}, \emph{w/o Agent-Infra} and \framework{} for the RAG-based Sales Agent and Tool-augmented Shopping Agent cases using
\texttt{gemini-2.5-flash}. The best score for each metric is shown in \textbf{bold}.}
\label{tab:app_diversity_gemini}
\end{table*}

\begin{table*}
\centering
\begin{tabular*}{\textwidth}{@{\extracolsep{\fill}}lcccc}
\toprule
\multirow{2}{*} & \multicolumn{2}{c}{GPT-5} & \multicolumn{2}{c}{Gemini-2.5-Flash}\\
\cmidrule(lr){2-3} \cmidrule(lr){4-5}
\textbf{Method} & \textbf{RAG-based} & \textbf{Tool-Augmented} & \textbf{RAG-based} & \textbf{Tool-Augmented}\\
\midrule
\emph{w/o ICP} & 27.4 (1.02) & 19.2 (1.31) & 22.6 (0.33) & 20.8 (0.75)\\
\emph{w/o Agent-Infra} & 14.6 (0.73) & 17.8 (0.92) & 16.0 (1.03) & 18.6 (2.01)\\
\framework{} & \best{28.8 (1.12)} & \best{24.0 (0.61)} & \best{27.8 (1.70)} & \best{24.2 (1.47)}\\
\bottomrule
\end{tabular*}
\caption{The average counts of unique bugs found over five runs with \emph{w/o ICP}, \emph{w/o Agent-Infra}, and \framework{} using \texttt{gpt-5} and \texttt{gemini-2.5-flash} as the user simulator models. The highest counts are shown in \textbf{bold}, standard deviations are in parenthesis.}
\label{tab:unique_bugs_close}
\end{table*}

\begin{table*}
\centering
\begin{tabular*}{\textwidth}{@{\extracolsep{\fill}}lccccccc}
\toprule
\textbf{Method} & \textbf{MTLD}$\uparrow$ & \textbf{HDD}$\uparrow$ & \textbf{logTTR}$\uparrow$ & \textbf{rootTTR}$\uparrow$ & \textbf{Vocab Size}$\uparrow$ & \textbf{Distinct@1/2/3}$\uparrow$ \\
\midrule
\multicolumn{7}{l}{\textbf{RAG-based Sales Agent}} \\
\midrule
w/o ICP       & 47.78 & 0.83 & 0.70 & 8.47 & 1630 & \best{0.07/0.24/0.37} \\
w/o Agent-Infra & 46.21 & 0.82 & 0.68 & 7.20 & 1415  & 0.05/0.21/0.33 \\
\framework{}  & \best{52.54} & \best{0.84} & \best{0.71} & \best{9.32} & \best{1937} & \best{0.07/0.24/0.37} \\
\midrule
\multicolumn{7}{l}{\textbf{Tool-Augmented Shopping Agent}}\\
\midrule
\emph{w/o ICP}       & 46.63 & \best{0.83} & 0.69 & 7.25 & 1168 & 0.08/0.27/0.44  \\
\emph{w/o Agent-Infra} & 49.42 & \best{0.83} & \best{0.70} & 7.45 & 1224 & 0.08/0.28/0.46 \\
\framework{}  & \best{49.93} & \best{0.83} & \best{0.70} & \best{7.58} & \best{1227} & \best{0.08/0.30/0.48} \\
\bottomrule
\end{tabular*}
\caption{Lexical diversity results of \emph{w/o ICP}, \emph{w/o Agent-Infra} and \framework{} for the RAG-based Sales Agent and Tool-augmented Shopping Agent cases using
\texttt{Qwen2.5-7B-Instruct}. The best score for each metric is shown in \textbf{bold}.}
\label{tab:app_diversity_qwen}
\end{table*}

\begin{table*}
\centering
\begin{tabular*}{\textwidth}{@{\extracolsep{\fill}}lccccccc}
\toprule
\textbf{Method} & \textbf{MTLD}$\uparrow$ & \textbf{HDD}$\uparrow$ & \textbf{logTTR}$\uparrow$ & \textbf{rootTTR}$\uparrow$ & \textbf{Vocab Size}$\uparrow$ & \textbf{Distinct@1/2/3}$\uparrow$ \\
\midrule
\multicolumn{7}{l}{\textbf{RAG-based Sales Agent}} \\
\midrule
w/o ICP       & 79.45 & 0.85 & \best{0.75} & 11.26 & 1297 & \best{0.14/0.49/0.71} \\
w/o Agent-Infra & 82.30 & 0.85 & 0.74 & 10.07 & 1324 & 0.11/0.41/0.63 \\
\framework{}  & \best{83.26} & \best{0.86} & \best{0.75} & \best{11.28} & \best{1340} & \best{0.14/0.49/0.71} \\
\midrule
\multicolumn{7}{l}{\textbf{Tool-Augmented Shopping Agent}}\\
\midrule
\emph{w/o ICP}       & 61.13 & \best{0.84} & \best{0.74} & 9.60 & 1076 & 0.13/0.47/0.69  \\
\emph{w/o Agent-Infra} & 63.24 & \best{0.84} & \best{0.74} & 9.79 & 1083 & \best{0.13/0.47/0.70} \\
\framework{}  & \best{63.62} & \best{0.84} & \best{0.74} & \best{9.84} & \best{1088} & \best{0.13/0.47/0.70} \\
\bottomrule
\end{tabular*}
\caption{Lexical diversity results of \emph{w/o ICP}, \emph{w/o Agent-Infra} and \framework{} for the RAG-based Sales Agent and Tool-augmented Shopping Agent cases using
\texttt{Llama-3.3-70B-Instruct}. The best score for each metric is shown in \textbf{bold}.}
\label{tab:app_diversity_llama}
\end{table*}

\begin{table*}
\centering
\begin{tabular*}{\textwidth}{@{\extracolsep{\fill}}lcccc}
\toprule
\multirow{2}{*} & \multicolumn{2}{c}{Qwen2.5-7B-Instruct} & \multicolumn{2}{c}{Llama-3.3-70B-Instruct}\\
\cmidrule(lr){2-3} \cmidrule(lr){4-5}
\textbf{Method} & \textbf{RAG-based} & \textbf{Tool-Augmented} & \textbf{RAG-based} & \textbf{Tool-Augmented}\\
\midrule
\emph{w/o ICP} & 16.2 (4.98) & 16.6 (2.05) & 20.6 (2.65) & 16.4 (3.20)\\
\emph{w/o Agent-Infra} & 9.8 (2.23) & 15.6 (1.85) & 14.6 (3.38) & 19.2 (3.97)\\
\framework{} & \best{20.6 (2.22)} & \best{21.2 (2.95)} & \best{23.2 (2.65)} & \best{21.2 (3.87)}\\
\bottomrule
\end{tabular*}
\caption{The average counts of unique bugs found over five runs with \emph{w/o ICP}, \emph{w/o Agent-Infra}, and \framework{} using \texttt{Qwen2.5-7B-Instruct} and \texttt{Llama-3.3-70B-Instruct} as the user simulator models. The highest counts are shown in \textbf{bold}, standard deviations are in parenthesis.}
\label{tab:unique_bugs_open}
\end{table*}

\section{Results Analysis}
\subsection{Automatic Evaluation}
\label{sec:app_auto_eval}
We use the \texttt{lexical\_diversity} package \cite{lexical-diver} to calculate the diversity score. Table~\ref{tab:app_diversity_gpt5} reports the results for interactions using \texttt{gpt-5} with the default hyperparameter settings. Table~\ref{tab:app_diversity_gemini} presents results for \texttt{gemini-2.5-flash} with \texttt{temperature=0.1}. Both models are called through APIs following their terms of use. 
Table~\ref{tab:app_diversity_qwen} and Table~\ref{tab:app_diversity_llama} present results for the open-source models \texttt{Qwen2.5-7B-Instruct}~\cite{qwen2.5} and \texttt{Llama-3.3-70B-Instruct}~\cite{llama3.3}, respectively, with \texttt{temperature=0.1}.
Across all models and use cases, \framework{} consistently achieves the best or comparable performance on most metrics.

Table~\ref{tab:unique_bugs_close} and Table~\ref{tab:unique_bugs_open} summarizes the average number of unique bugs identified over five runs for interactions generated by commercial and open-source models respectively. Consistent with our previous results, \framework{} uncovers the highest number of unique errors, followed by the \emph{w/o ICP} setting. Removing bottom-up knowledge leads to the sharpest drop in error identification. These findings demonstrate that \framework{} is robust across different models.

\begin{table*}[t]
    \centering
    \small
    \begin{tabular*}{\textwidth}{@{\extracolsep{\fill}} p{0.3\textwidth} p{0.3\textwidth} p{0.3\textwidth}}
    \toprule
    \textbf{w/o ICP} & \textbf{\framework{}} &\textbf{w/o Agent-Infra}\\
    \midrule
    \textbf{User}: \textcolor{magenta}{I'm interested in the \hlc[SkyBlue]{Product\_A} delivery robot. How does it compare to other delivery solutions?} \newline
    \textbf{Agent}: The Product\_A delivery robot is designed for efficient indoor deliveries, thriving in dynamic environments like restaurants and cafes. It uses smart obstacle avoidance and can interact with elevators for secure deliveries. Compared to other solutions, Product\_A offers cost-effectiveness and reliability, reducing labor costs and enhancing service speed. For more details, you can explore our website. Which model(s) are you interested in? \newline
    \textbf{User}: \textcolor{magenta}{I'm mainly interested in delivery solutions and comparing Product\_A. Can you tell me more about its benefits over other robots?} \newline
    \textbf{Agent}: The Product\_A delivery robot excels in indoor environments like restaurants and cafes, offering smart obstacle avoidance and elevator interaction for secure deliveries. Compared to other robots, Product\_A is cost-effective, reliable, and enhances service speed, reducing labor costs. \newline
    \textbf{User}: Can it handle \hlc[SkyBlue]{different types of deliveries} efficiently? \newline
    \textbf{Agent}: Yes, it can efficiently handle various types of deliveries. It excels in indoor environments, using smart navigation and obstacle avoidance to ensure timely deliveries. This enhances service speed and reduces labor costs, making it a versatile solution for businesses. To speed up your process, would you like to provide your business name? \newline
    \textbf{User}: I'm currently exploring options. Thanks for the info!
    &
    \textbf{user}: hey, how does \hlc[SkyBlue]{Product\_A} compare with other delivery robots? \hlc[SpringGreen]{need it for agriculture operations}. \newline
    \textbf{Agent}: Product\_A is designed for efficient delivery operations. It enhances operational efficiency and optimizes labor resources. Which model(s) are you interested in? \newline
    \textbf{User}: i'm trying to evaluate if Product\_A is suitable for \hlc[SpringGreen]{outdoor environments in farming}. what's its durability like? \newline
    \textbf{Agent}: \textcolor{orange}{Product\_A is designed for robust outdoor use, making it suitable for agricultural operations.} It is built to withstand various environmental conditions, ensuring durability and reliability in farming tasks. It's battery life varies based on usage, but it typically runs for several hours before needing a recharge. For more specific details, please visit our website. Do you want to rent or buy our robots? \newline
    \textbf{User}: \hlc[SkyBlue]{how long does it take to charge?} i'm considering all options. \newline
    \textbf{Agent}: It takes approximately \textcolor{orange}{3 hours} to fully charge from 0 to 100\%. This efficient charging time ensures minimal downtime for your agricultural operations. For more details, please visit the our website. To expedite your rental process, could you please provide your business name? \newline
    \textbf{User}: \hlc[SpringGreen]{it's AgriFuture Technologies}. not renting yet, just exploring options.
    &
    \textbf{User}: \textcolor{magenta}{How do the Product\_B and Product\_C compare} \hlc[SpringGreen]{for farm work}? \newline
    \textbf{Agent}: The Product\_B is versatile and can be used for multiple purposes, including making tea, coffee, and cocktails. However, it is not specifically designed for farm work. The Product\_C is a delivery robot, primarily intended for transporting items, which might not be suitable for farm tasks. If you have specific requirements for farm work, please let me know, and I can provide more tailored information! \newline
    \textbf{User}: What about Product\_A robot? \newline
    \textbf{Agent}: The Product\_A robot is a delivery robot designed for transporting items, primarily in business settings. It enhances operational efficiency and optimizes labor resources. For more details, visit our website. We have amazing worker, delivery, cleaning, and multipurpose robots. Which model(s) are you interested in? \newline
    \textbf{User}: Delivery bot like Product\_A I think. \hlc[SpringGreen]{Can it handle outdoor conditions?} \newline
    \textbf{Agent}: Currently, our products are primarily designed for indoor business environments, but they may also be suitable for outdoor conditions, such as agricultural settings. If you have specific requirements for outdoor use, please contact our customer service team for more details. \newline
    \textbf{User}: Got it. Thanks for the info. \\
    \bottomrule
    \end{tabular*}
    \caption{Full interaction samples from our human evaluation under three conditions: \framework{} (middle), \emph{w/o ICP} (left), and \emph{w/o Agent-Infra} (right) with the same user's goal: compare products. Text highlighted in \hlc[SkyBlue]{blue} reflects generated content tied to infrastructure knowledge, while text in \hlc[SpringGreen]{green} reflects content consistent with ICP attributes. Agent responses marked in \textcolor{orange}{orange} indicate faithfulness errors. On the left, user queries in \textcolor{magenta}{magenta} indicate repetitive and generic responses. On the right, the user queries in \textcolor{magenta}{magenta} are ones which are irrelevant to product attributes. Product\_A, Product\_B and Product\_C are placeholders used to mask the actual product names. }
    \label{tab:error_examples_full}
\end{table*}

\subsection{The Agent Errors Analysis}
\begin{wrapfigure}{*}{0.9\columnwidth}
    \centering
    \vspace{-1em}
    \includegraphics[width=\columnwidth]{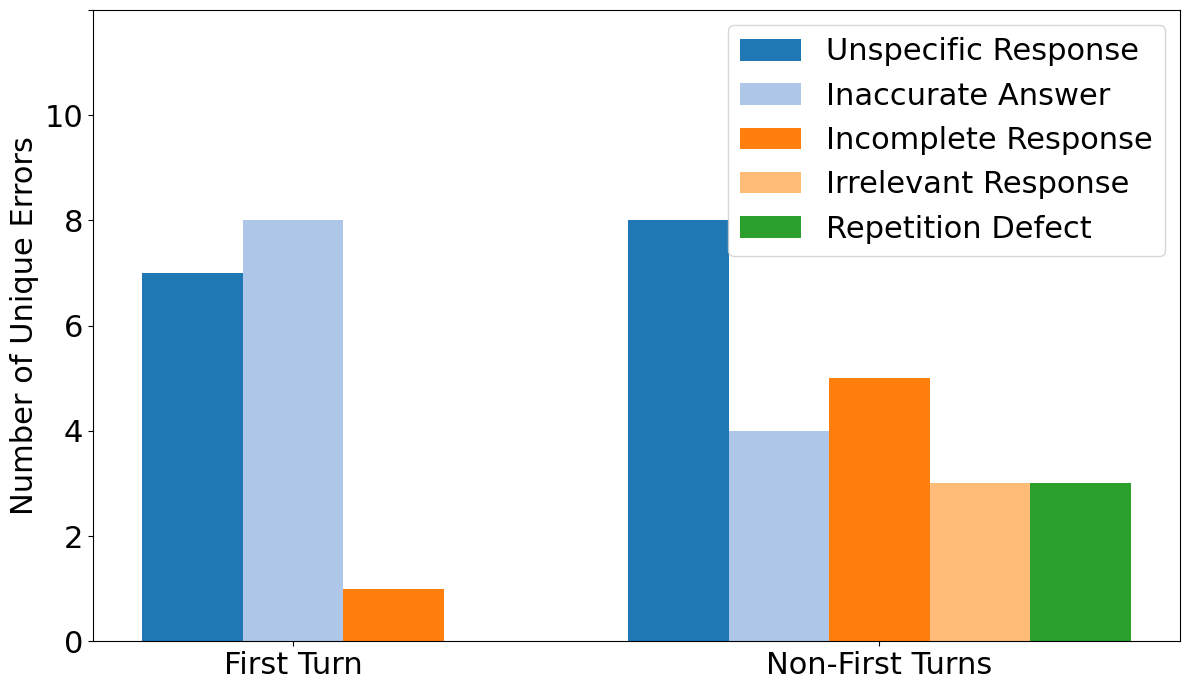}
    \caption{The number of unique errors in each category, grouped by their occurrence in the first turn versus subsequent turns.}
    \vspace{-3em}
    \label{fig:failure_turn}
\end{wrapfigure}

Figure \ref{fig:failure_turn} shows the distribution of high-level error categories identified by \framework{}, along with when these errors occurred within the interaction. 
We observe that most errors occur after the first turn, particularly "Incomplete Response", "Irrelevant Response" and "Repetition Defect", which mostly emerge in non-first turns. This highlights the importance of multi-turn evaluation for reliably assessing agent behavior. This finding is consistent with prior work showing that LLMs struggle to stay on track in extended dialogues and have difficulty recovering once errors occur \cite{laban2025llmslostmultiturnconversation}.
Table \ref{tab:sample_errors} shows more interaction examples for each high-level error category.

\begin{table*}[t]
    \centering
    \small
    \begin{tabularx}{\textwidth}{X}
    \toprule
      \arrayrulecolor{gray!20}\specialrule{0pt}{0pt}{-2pt}
      \rowcolor{gray!20} \normalsize\textbf{Unspecific Response}\\
      \arrayrulecolor{black}
    \toprule
      \textbf{Bug Description}: Did not provide the specific weight capacity for Product\_A when requested \\
      \midrule
      \textbf{Example}: \\
      \textit{[Earlier parts omitted for brevity]} \\
      \textbf{User}: hmm. what about weight? Can it carry heavy items or no? \\
      \textbf{Agent}: The Product\_A is designed to handle a variety of delivery tasks, but it has weight limitations. For specific weight capacity details, please refer to the product specifications on our website or contact our customer service for more information. \\
      \textbf{User}: ugh, don't have time for that. can't you just give me a number here? \\
      \textbf{Agent}: The Product\_A is capable of handling various delivery tasks, but it does have specific weight limitations. \textcolor{orange}{For detailed information on its weight capacity, please check the product specifications on our website or contact our customer service.}\\

    \toprule
      \arrayrulecolor{gray!20}\specialrule{0pt}{0pt}{-2pt}
      \rowcolor{gray!20}\normalsize\textbf{Inaccurate Answer} \\
      \arrayrulecolor{black}
    \toprule
      \textbf{Bug Description}: Incorrectly claimed Company\_name ships internationally \\
      \midrule
      \textbf{Example}: \\
      \textbf{User}: Hey, do you guys offer international shipping? \\
      \textbf{Agent}: \textcolor{orange}{Yes, we offer international shipping for our robots. Delivery times may vary depending on the destination.} If you have any specific questions or need further assistance, feel free to ask! \\
      \textbf{User}: Wait, are you sure? I thought it was different. \\
      \textbf{Agent}: \textcolor{orange}{I apologize for the confusion earlier. Currently, Company\_name only operates within the US and does not offer international shipping.} If you have any other questions or need further assistance, feel free to ask! \\
      \textbf{User}: That's disappointing. Can you be more consistent next time? \\
      
    \toprule
      \arrayrulecolor{gray!20}\specialrule{0pt}{0pt}{-2pt}
      \rowcolor{gray!20}\normalsize\textbf{Incomplete Response} \\
      \arrayrulecolor{black}
    \toprule
      \textbf{Bug Description}: Did not provide requested product comparison (e.g., Product\_C vs similar robots) \\
      \midrule
      \textbf{Example}: \\
      \textbf{User}: Can you tell me how Product\_C compares with similar delivery robots? \\
      \textbf{Agent}: The Product\_C delivery robot is designed for efficient and reliable delivery services in business environments. It offers advanced navigation and obstacle avoidance, making it suitable for various settings. \textcolor{orange}{While I can't provide a direct comparison with other specific delivery robots, Product\_C is known for its robust performance and adaptability.} If you have more specific questions about its features or capabilities, feel free to ask! \\
      \textbf{User}: I asked about a comparison, not just more features... Can you please be more useful? \\

    \toprule
      \arrayrulecolor{gray!20}\specialrule{0pt}{0pt}{-2pt}
      \rowcolor{gray!20}\normalsize\textbf{Irrelevant Response} \\
      \arrayrulecolor{black}
    \toprule
      \textbf{Bug Description}: Requested personal contact information and/or created support tickets without user consent or clear necessity \\
      \midrule
      \textbf{Example}: \\
      \textbf{User}: What's the cleaning efficiency of the Product\_D? And how does it navigate? \\
      \textbf{Agent}: The Product\_D is designed for commercial cleaning with high efficiency, utilizing advanced sensors and mapping technology for precise navigation. If you have more questions or need further assistance, feel free to ask! \\
      \textbf{User}: This isn't what I asked for. Can you provide more details on these sensors and its exact cleaning efficiency? \\
      \textbf{Agent}: The Product\_D cleans at a rate of 21,958 sqft/hr. It uses a combination of sensors, including a monocular panoramic camera, binocular camera, LiDAR, 3D ToF camera, ultrasonic wave, IMU, and electronic bumper for navigation. For more details, you can visit the Company\_name website. \textcolor{orange}{I'm sorry to hear that! We are creating a support ticket now. Could you provide me with your email address and phone number?} \\

    \toprule
      \arrayrulecolor{gray!20}\specialrule{0pt}{0pt}{-2pt}
      \rowcolor{gray!20}\normalsize\textbf{Repetition Defect} \\
      \arrayrulecolor{black}
    \toprule
      \textbf{Bug Description}: Repeated the same non-answer across turns despite user frustration \\
      \midrule
      \textbf{Example}: \\
      \textbf{User}: hey, can u tell me the shipment status with the tracking number i got? \\
      \textbf{Agent}: I'm sorry, but I can't access shipment details directly. Please use the tracking number you received via email to check your order status online. If you need further assistance, feel free to ask! \\
      \textbf{User}: Uhh, yeah I got the tracking number in the email, just need some details about the current status! \\
      \textbf{Agent}: \textcolor{orange}{I'm sorry, but I can't access shipment details directly. Please use the tracking number you received via email to check your order status online. If you need further assistance, feel free to ask!} \\
      \textit{[Later interaction repeat the same answer multiple times]} \\
    \bottomrule
    \end{tabularx}
    \caption{The synthesized interaction examples generated by the \framework{} framework. The sentences highlighted in \textcolor{orange}{orange} represent the agent errors for the corresponding error category.}
    \label{tab:sample_errors}
\end{table*}

\clearpage
\section{Prompts}
\subsection{Profile Construction}
\label{sec:prompt_profile_construction}
\begin{tcolorbox}[
  breakable,
  enhanced,
  colback=gray!5!white,
  colframe=gray!60!black,
  title=Augment attributes with the agent infrastructure knowledge,
]
\small{
Given the agent infrastructure knowledge, your task is to generate \{num\_personas\} set of personal attributes for the new categories.\\
\\
Examples: \\
\#\#\#\#\#\#\#\#\#\#\#\#\#\#\#\#\#\#\#\#\#\#\#\# \\
Generate 2 sets of attributes. \\
agent infrastructure knowledge:  \\
\{knowledge\}\\
\\
Here are new attribute descriptions that you need to generate values for: \\
business\_type: The type of business or industry the individual is associated with, such as technology, finance, healthcare, retail, or manufacturing. \\
company\_name: The name of the company or organization the individual is associated with. \\
job\_information: Detailed information about the individual's professional roles, including their official job title, primary responsibilities, and total years of relevant work experience within the industry or field. Such as marketing manager with 7 years of experience in the industry, taking care of exploration and expansion of the market. \\
company\_size: The size of the company or organization the individual is associated with, better with the number of employees. \\
\\
Return: \\
\{\{"attributes": [\{\{"business\_type": "commercial cleaning services", "company\_name": "House Cleaner", "job\_information": "Facilities Manager with 8 years of experience in commercial cleaning services, managing cleaning staff and implementing automation solutions", "company\_size": "100-200 employees"\}\}, \{\{"business\_type": "hospitality (hotels, bars, restaurants)", "company\_name": "Lily Motel", "job\_information": "Food \& Beverage Director with 10 years of experience in hospitality, overseeing bar and restaurant operations and technology adoption", "company\_size": "11-50 employees"\}\}]\}\} \\
\\
Rules: \\
- You should return a JSON list of \{num\_personas\} sets of attributes values. \\
- The result should be diverse and not repetitive. \\
- The attributes values should be aligned with the corporate overview. \\

\#\#\#\#\#\#\#\#\#\#\#\#\#\#\#\#\#\#\#\#\#\#\#\# \\
Generate \{num\_personas\} sets of attributes. \\
\\
agent infrastructure knowledge: \\
\{knowledge\} \\
\\
Here are the categories that you need to generate values for, one value per category: \\
\{category\_description\} \\
\\
Return: \\
}
\end{tcolorbox}

\begin{tcolorbox}[
  breakable,
  enhanced,
  colback=gray!5!white,
  colframe=gray!60!black,
  title=Validate attributes,
]
\small{
Given a set of user attributes, your task is to check whether each pair of attributes is consistent. You should first generate the thought and then return the response "Yes" or "No" in JSON FORMAT. \\
\\
Example: \\
\#\#\#\#\#\#\#\#\#\#\#\#\#\#\#\#\#\#\#\#\#\#\#\# \\
User attributes: \\
job\_information: operations manager with 10 years of experience in hospitality, overseeing automation and service efficiency \\
business\_type: commercial cleaning services \\
company\_size: 11-50 employees \\
company\_name: Caesars Entertainment \\
\\
Return: \\
\{\{ \\
"thought": "a commercial cleaning services company will not name as Entertainment company", \\
"response": "No" \\
\}\}
\\
\#\#\#\#\#\#\#\#\#\#\#\#\#\#\#\#\#\#\#\#\#\#\#\# \\
User attributes: \\
customer\_type: returning customer \\
deal\_stage: new leads \\
\\
Return: \\
\{\{ \\
\quad    "thought": "a returning customer will not be a new leads", \\
\quad    "response": "No" \\
\}\} \\
\\
\#\#\#\#\#\#\#\#\#\#\#\#\#\#\#\#\#\#\#\#\#\#\#\# \\
User attributes: \\
\{user\_attributes\} \\
\\
Return: \\
}
\end{tcolorbox}

\begin{tcolorbox}[
  breakable,
  enhanced,
  colback=gray!5!white,
  colframe=gray!60!black,
  title=Convert user attributes to a user profile,
]
\small{
Given a list of user attributes, your task is to convert them into a natural language description of a user profile. \\
\\
Rules:\\
- Write the response in the second person. \\
- Begin the profile with: "You are <name> who is <TRAIT 1, ..., TRAIT 5>.", where [TRAIT 1, ..., TRAIT 5] represents the assigned Big Five personality traits. \\
- Do not add any prefixes such as "Profile:" or "Customer Profile:". \\
\\
User attributes: \\
\{user\_attr\} \\
\\
Profile: \\
}
\end{tcolorbox}

\onecolumn
\subsection{User simulator prompt}
\label{sec:prompt_convo_gen}
\begin{tcolorbox}[
  breakable,
  enhanced,
  colback=gray!5!white,
  colframe=gray!60!black,
  title=User simulator's system prompt,
]
\small{
You are a customer interacting with an agent. The agent is supplied by the following company: \\
\{summary\} \\
\{profile\} \\
You have the following goal when interacting with this agent:
\{goal\} \\
Here is the content that you might be interested in and might have questions about: \\
\{knowledge\} \\
\\
Rules: \\
- Just generate one line at a time to simulate the user's message. Keep the response brief and concise. \\
- Do not give away all the instruction at once. Only provide the information that is necessary for the current step. \\
- Do not hallucinate information that is not provided in the instruction. For example, if the agent asks for the order id but it is not mentioned in the instruction, do not make up an order id, just say you do not remember or have it. \\
- If the instruction goal is satisified, generate '\#\#\#STOP\#\#\#' as a standalone message without anything else to end the conversation. \\
- Do not repeat the exact instruction in the conversation. Instead, use your own words to convey the same information. \\
- Try to personalize your question to the your background, such as job information, household, event or activities you did or going to do. \\
- Adjust your tone/style based on your personalities. \\
- As a customer, you tend to write short questions with occasional typos or incomplete sentences. Replicate the chitchat typing behavior of a human customer and begin the conversation with a question to achieve your goal. \\
}
\end{tcolorbox}

\subsection{Agent Evaluation Metrics}
\label{sec:prompt_agent_eval}
All the agent evaluation prompts start with the \textbf{TASK\_INSTRUCTION} followed by the specific rubrics for each metric. 

\begin{tcolorbox}[
  enhanced,
  colback=gray!5!white,
  colframe=gray!60!black,
  title=TASK\_INSTRUCTION
]
\small{
  Given a conversation between a user and an AI assistant, your task is to evaluate the last message from the AI assistant according to the criteria below, assigning a score from 0 to 4. Follow the detailed descriptions to guide your assessment. Return the score and reasoning in JSON FORMAT. \\
  \\
  Format: \\
\{\{ \\
\quad    "score": <integer 0-4>, \\
\quad    "reason": "<reasoning>" \\
\}\} \\
}
\end{tcolorbox}

\begin{tcolorbox}[
  enhanced,
  colback=gray!5!white,
  colframe=gray!60!black,
  title=Helpfulness
]
\small{
[TASK\_INSTRUCTION] \\
Helpfulness: \\
Description: Determine the effectiveness and value of the AI assistant's responses in the last message in addressing the user's needs. \\
Evaluation Criteria: \\
• 4 - The response is extremely helpful and completely aligned with the spirit of what the prompt was asking for. \\
• 3 - The response is mostly helpful and mainly aligned with what the user was looking for, but there is still some room for improvement.  \\
• 2 - The response is partially helpful but misses the overall goal of the user's query/input in some way. The response did not fully satisfy what the user was looking for.  \\
• 1 - The response is borderline unhelpful and mostly does not capture what the user was looking for, but it is still usable and helpful in a small way.  \\
• 0 - The response is not useful or helpful at all. The response completely missed the essence of what the user wanted.\\
\\
Here is the conversation between user and AI assistant: \\
\{conversation\} \\
\\
Evaluation Form: \\
}
\end{tcolorbox}

\begin{tcolorbox}[
  enhanced,
  colback=gray!5!white,
  colframe=gray!60!black,
  title=Coherence
]
\small{
[TASK\_INSTRUCTION] \\
Coherence: \\
Description: Determine the logical flow and consistency of the AI assistant's responses in the last message. \\
Evaluation Criteria: \\
• 4 (Perfectly Coherent and Clear) - The response is perfectly clear and self-consistent throughout. There are no contradictory assertions or statements, the writing flows logically and following the train of thought/story is not challenging. \\
• 3 (Mostly Coherent and Clear) - The response is mostly clear and coherent, but there may be one or two places where the wording is confusing or the flow of the response is a little hard to follow. Over all, the response can mostly be followed with a little room for improvement.  \\
• 2 (A Little Unclear and/or Incoherent) - The response is a little unclear. There are some inconsistencies or contradictions, run on sentences, confusing statements, or hard to follow sections of the response.  \\
• 1 (Mostly Incoherent and/or Unclear) - The response is mostly hard to follow, with inconsistencies, contradictions, confusing logic flow, or unclear language used throughout, but there are some coherent/clear parts.  \\
• 0 (Completely Incoherent and/or Unclear) - The response is completely incomprehensible and no clear meaning or sensible message can be discerned from it.\\
\\
Here is the conversation between user and AI assistant: \\
\{conversation\} \\
\\
Evaluation Form: \\
}
\end{tcolorbox}

\begin{tcolorbox}[
  enhanced,
  colback=gray!5!white,
  colframe=gray!60!black,
  title=Verbosity
]
\small{
[TASK\_INSTRUCTION] \\
Verbosity: \\
Description: How concise is the last message from the AI assistant. Does it use more words than needed? \\
Evaluation Criteria: \\
• 4 (Verbose) - The response is particularly lengthy, wordy, and/or extensive with extra details given what the prompt requested from the assistant model. The response can be verbose regardless of if the length is due to repetition and incoherency or if it is due to rich and insightful detail. \\
• 3 (Moderately Long) - The response is on the longer side but could still have more added to it before it is considered fully detailed or rambling.  \\
• 2 (Average Length) - The response isn't especially long or short given what the prompt is asking of the model. The length is adequate for conveying a full response but isn't particularly wordy nor particularly concise.  \\
• 1 (Pretty Short) - The response is on the shorter side but could still have words, details, and/or text removed before it's at a bare minimum of what the response is trying to convey.  \\
• 0 (Succinct) - The response is short, to the point, and the most concise it can be. No additional information is provided outside of what is requested by the prompt (regardless of if the information or response itself is incorrect, hallucinated, or misleading. A response that gives an incorrect answer can still be succinct.).\\
\\
Here is the conversation between user and AI assistant: \\
\{conversation\} \\
\\
Evaluation Form: \\
}
\end{tcolorbox}

\begin{tcolorbox}[
  enhanced,
  colback=gray!5!white,
  colframe=gray!60!black,
  title=Relevance
]
\small{
[TASK\_INSTRUCTION] \\
Relevance: \\
Description: How much the last message from the AI assistant is addressing the question/input from the customer and to what degree? \\
Evaluation Criteria: \\
• 4 (Very Relevant) - The response is addressing the input question in its entirety, and just the question. \\
• 3 (Additional information) - The response is fully addressing the question, but also providing information not directly related to the question alongside.  \\
• 2 (Partially Relevant) - The response is partially addressing the question i.e., only a part of the question is addressed.  \\
• 1 (Pretty Short) - The response only address small portion of the question.  \\
• 0 (Not Relevant) - The response is not relevant to the user's query/input at all.\\
\\
Here is the conversation between user and AI assistant: \\
\{conversation\} \\
\\
Evaluation Form: \\
}
\end{tcolorbox}

\begin{tcolorbox}[
  enhanced,
  colback=gray!5!white,
  colframe=gray!60!black,
  title=Faithfulness
]
\small{
[TASK\_INSTRUCTION] \\
Faithfulness: \\
Description: How accurate/factual the answer seems to be, based on: provided knowledge content and majorly agreed common knowledge. \\
Evaluation Criteria: \\
• 4 - The response is completely correct and faithfulness to the knowledge with no necessary details missing and without false, misleading, or hallucinated information. If the prompt asks the assistant to do a task, the task is completely done and addressed in the response.  \\
• 3 - The response is mostly correct and faithfulness to the knowledge with a small amount of missing information. It contains no misleading information or hallucinations. If the prompt asks the assistant to perform a task, the task is mostly successfully attempted.  \\
• 2 - The response contains a mix of correct and incorrect information. The response may miss some details, contain misleading information, or minor hallucinations, but is more or less aligned with what the prompt asks for. If the prompt asks the assistant to perform a task, the task is attempted with moderate success but still has clear room for improvement.  \\
• 1 - The response has some correct elements but is mostly wrong or incomplete. The response may contain multiple instances of hallucinations, false information, misleading information, or irrelevant information. If the prompt asks the assistant to do a task, the task was attempted with a small amount of success.   \\
• 0 - The response is completely incorrect. All information provided is wrong, false or hallucinated. If the prompt asks the assistant to do a task, the task is not at all attempted, or the wrong task was attempted in the response. The response is completely irrelevant to the prompt.\\
\\
Here is the knowledge that assistant's response should be grounded: \\
\{knowledge\} \\
\\
Here is the conversation between user and AI assistant: \\
\{conversation\} \\
\\
Evaluation Form: \\
}
\end{tcolorbox}

\subsection{Identify Errors}
\label{sec:prompt_identify_bugs}
In order to identify more specific agent errors, we adopt two-stage process. In the first stage, we prompt an LLM to generate high level failure categories using the problematic turns and their score justifications. In the second stage, the output from the first stage along with the justifications are given to an LLM to generate fine-grained, unique bug statement. 

\begin{tcolorbox}[
  enhanced,
  colback=gray!5!white,
  colframe=gray!60!black,
  title=identify high-level error categories
]
\small{
Given the failure cases from user-agent interactions, your task is to categorize them into high-level failure categories. \\
\\
Rules: \\
- For each case, first provide a short reasoning statement (“thought”) explaining why the case belongs to a particular category. \\
- Assign a concise and meaningful name to the high-level failure category. \\
- Provide a clear description that characterizes the nature of this category. \\
- Return the final output strictly in JSON format.\\
\\
Format: \\
\{\{"errors": [\{\{"thought": "<brief reasoning for the categorization>", "high\_level\_error\_category": "<name of the category>", "description": "<definition of the category>"\}\}]\}\}\\
\\
Failure cases:\\
\{items\}\\
\\
Return:\\
}
\end{tcolorbox}

\begin{tcolorbox}[
  enhanced,
  colback=gray!5!white,
  colframe=gray!60!black,
  title=identify unique errors
]
\small{
Given the failure cases from user-agent interactions, and the high-level failure categories, your task is to deduplicate these failure cases and identity fine-grained unique errors under each high-level failure category. \\
\\
Rules: \\
- Do not follow the error types from "helpfulness", "coherence", "verbosity", "faithfulness", "relevance" anymore. Instead, concentrate specifically on the agent’s behavior. For example, identify which specific pieces of information are missing. Each missing piece of information should be treated as a separate unique error. \\
- Return the error types in the JSON format, where high\_level\_error\_category is one of the high-level error categories from the choice list, unique\_error\_description is the description of the unique error, example is item\_id corresponding to that unique error. \\
\\
Format: \\
\{\{"unique\_errors": [\{\{"high\_level\_error\_category": "<high-level error category name>", "unique\_error\_description": "<unique\_error\_description>", "example": ["<item\_id1>, <item\_id2>, ..."]\}\}]\}\}\\
\\
High-level failure categories: \\
\{high\_level\_failure\_categories\} \\
\\
Failure cases:\\
\{items\}\\
\\
Return:\\
}
\end{tcolorbox}

\clearpage
\twocolumn


\end{document}